\pdfoutput=1

\documentclass[11pt]{article}

\usepackage{acl}

\usepackage{times}
\usepackage{latexsym}

\usepackage[T1]{fontenc}

\usepackage[utf8]{inputenc}

\usepackage{microtype}

\usepackage{inconsolata}

\usepackage{graphicx}

\usepackage{enumitem}
\usepackage{booktabs}
\usepackage[english]{babel}
\usepackage[english=british]{csquotes}

\title{Classifier identification in Ancient Egyptian\\as a low-resource sequence-labelling task}

\author{Dmitry Nikolaev$^1$ \quad Jorke Grotenhuis$^2$ \quad Haleli Harel$^2$ \quad Orly Goldwasser$^2$ \\
        $^1$University of Manchester \quad $^2$Hebrew University of Jerusalem \\
        \texttt{dmitry.nikolaev@manchester.ac.uk},\\\texttt{\{jorke.grotenhuis,haleli.harel,orly.goldwasser\}@mail.huji.ac.il}}

\begin{document}

\maketitle

\begin{abstract}
The complex Ancient Egyptian (AE) writing system was characterised by widespread use of graphemic classifiers 
(determinatives):
silent (unpronounced) hieroglyphic signs clarifying the meaning or indicating the pronunciation of the host word. 
The study of classifiers has intensified in recent years with the launch 
and quick growth of the iClassifier project, a web-based platform for annotation and analysis of 
classifiers in ancient and modern languages. Thanks to the data contributed by the project 
participants, it is now possible to formulate the identification of classifiers in AE texts as an NLP task. 
In this paper, we make first steps towards solving this task by implementing a series of sequence-labelling neural 
models, which achieve promising performance despite the modest amount of training data. We discuss tokenisation and 
operationalisation issues arising from tackling AE texts and contrast our approach with frequency-based baselines.
\end{abstract}

\section{Introduction}

The Ancient Egyptian language and writing system, which belong to the earliest stratum of intangible cultural heritage
available to researchers, possess a range of interesting features. One of them is widespread use of classificatory
signs, called \textit{determinatives} in earlier literature. These classifiers (hereafter CLFs in ambiguous contexts,
in order to avoid confusion with classifier models) are hieroglyphic signs attached, singly or in combinations, 
to words of different parts of speech and used mostly to highlight some aspect of the host word's meaning or
pronunciation \citep{goldwasser2023guide,goldwasser2012determinatives}.
Egyptian graphemic classifiers are usually understood to be a purely written phenomenon,
i.e., unlike classifiers in contemporary spoken languages \citep{grinevald2015clfs}, they were not pronounced.
Classifiers of this type have been most intensively studied in Ancient Egyptian, but they have been also
described in Sumerian \citep{selz2017sumerian} and Luwian \citep{payne2017determination}, and it is argued
that the ancient Chinese writing system was built on similar principles \citep{goldwasserhandel2024chinese}.

The computational research on the Ancient Egyptian language is in its infancy. A~comprehensive overview
of studies of ancient languages utilising machine-learning methods, prepared by \citet{sommerschield2023survey},
mentions only a couple of works on Egyptian, and all of them deal with technical tasks, such as
optical character recognition and spectrography-based dating. Neither do we know of any computational
works tackling classifiers/determinatives in other ancient scripts.

At the same time, the field of classifier studies has been progressing rapidly in recent years. To a large extent
this is due to the launch of iClassifier \citep{hareletal2024iclassifier}, a dedicated platform for analysis
of classifiers in ancient and spoken languages, which ensures comparability between annotated corpora.
By providing such a platform, the project aims to facilitate both the study of individual classification traditions
and, by means of semantic annotations with CONCEPTICON labels \citep{concepticon}, cross-cultural analyses 
of classification systems.

The particular structure of any given corpus is dependent on its creator, and the project includes resources of
two basic types:
\begin{enumerate}
    \item Full-text corpora, which include annotations for both classified and unclassified wordforms from a particular text or set of texts.
    \item Topical corpora, which include data points of a particular type, e.g., lexical borrowings or items from a 
    particular lexical class.
\end{enumerate}
Corpora of the first type are more informative, but in practice they presuppose the existence of already-digitised
texts that can be imported in iClassifier wholesale and then annotated. In some cases, the target texts have not
yet been digitised, and only words or phrases of particular interest are manually entered.

Work on projects of both types could be facilitated by the existence of a trained classifier model, which would highlight
potential CLF tokens in inputs. If such a classifier attains a high degree of accuracy, it will then be possible
to conduct fast analyses of large digitised textual corpora, which have been published for, e.g.,
Ancient Egyptian \citep{tla}, Sumerian,\footnote{\url{https://etcsl.orinst.ox.ac.uk/}}
Luwian,\footnote{\url{http://web-corpora.net/LuwianCorpus/search/}} and ancient Chinese \citep{xu2024semantic}.
From the research perspective, an accurate discriminative classifier model will serve as a first step towards 
building a more interpretable generative model for word classification in ancient complex scripts and spoken languages.

In this study, we take first steps towards developing such a classifier on the basis of the Coffin Texts
corpus, as of today the largest annotated full-text corpus in the iClassifier system.

\section{Data}
\label{sec:data}

\subsection{The corpus}
\label{ssec:corpus}

The main dataset used in this study is a subset of the so-called Coffin Texts \citep{buck1935coffin},
a collection of spells painted on burial coffins of the First Intermediate period (c.~2130--1938 BCE)
and the Middle Kingdom (1938~-- c.~1630 BCE). A~subset of the spells forms one of full-text projects in iClassifier,
i.e.~it includes both classified and unclassified data points in the proportions reflecting the linguistic usage
of the time, which makes it suitable for training a classifier-identification model. The corpus is word based:
individual data points are wordforms, which is the standard annotation practice for ancient
texts in iClassifier.\footnote{Modern languages usually need sentences as data points, while the ancient
Chinese corpora, conversely, decompose individual signs into the phonetic and semantic component and treat the
latter as a classifier. See \citet{xu2024semantic} for details.}

\begin{figure}
    \centering
    \includegraphics[width=0.9\linewidth]{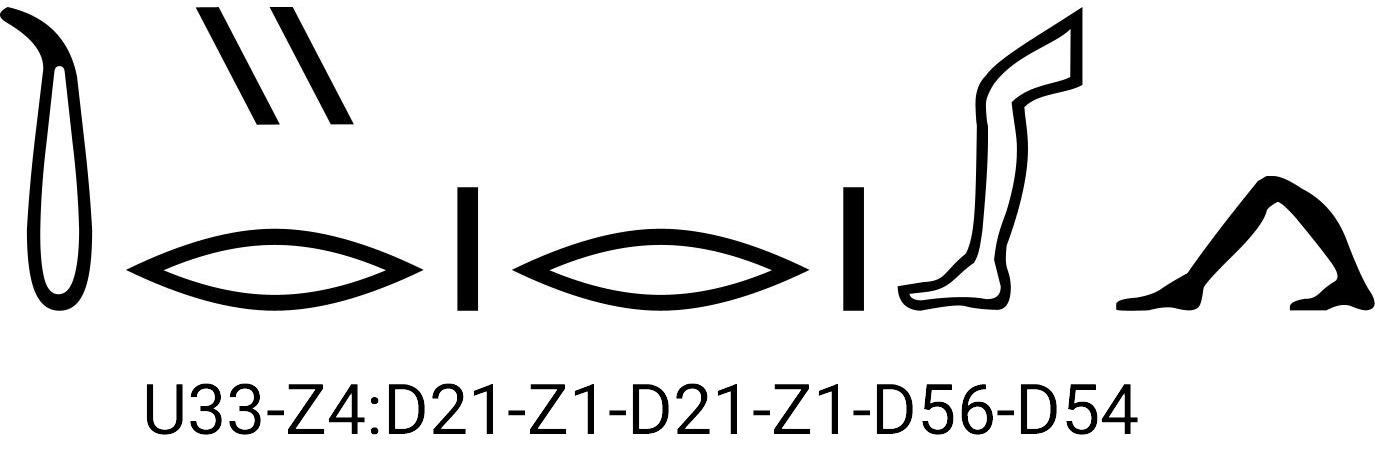}
    \caption{A form of the verb \textit{trr} `to race' represented in hieroglyphs and in the Manuel de Codage
    transcription. The last two signs are unpronounced semantic classifiers putting `race' in the [MOVEMENT] category.}
    \label{fig:mdc-example}
\end{figure}

This corpus, similarly to other corpora in the project, relies on a broad definition of the term 
\textit{classifier} that encompasses not only semantic CLFs\footnote{Including so-called \enquote{repeater CLFs},
where an unpronounced pictorial logogram expresses the same meaning as that conveyed by a phonologically-encoded word.}
but also phonograms presenting redundant phonological information, such as phono-repeaters.
These sign functions can be tagged in the UI as \enquote{semantic classifiers} and \enquote{phonetic 
classifiers}, respectively. Additional tagged signs pertain to common \enquote{grammatical classifiers}, which represent 
the number or gender of the host word. 
As a first step we do not distinguish between different CLF types but try and identify all non-autonomous 
signs \citep[157]{polis2015sign}.

The fully-annotated subset of the Coffin Texts corpus contains 74106 data points. However, many wordforms are repeated several
times with the same CLFs, which reduces the effective size of the dataset to 8423 types, randomly split 
into 6739 train, 842 development, and 842 test data points. Table~\ref{tab:clf-counts} shows the statistics
of the number of CLFs per data point.

The setting therefore can be characterised as extremely
low resource since not only the dataset itself is small, but there are no language models pre-trained on the target
language.\footnote{The most closely related language with a sizeable corpus is Coptic, which was written in
an alphabetic script and presents a tough low-resource scenario in itself, cf., e.g.,
\citet{gessler-zeldes-2022-microbert}.}

We also use a small (404 data points) corpus of wordforms from Late Egyptian
narratives\footnote{\url{https://thesaurus-linguae-aegyptiae.de/text/MTBRL3MIIJDKXAOF2336WRLMZA}}
as a separate out-of-domain test set. This smaller corpus represents a different textual genre, a folktale, 
and was compiled later, in the 13th century BCE, compared to the Coffin Texts, which are dated to
22nd--17th c.\ BCE.

\subsection{The transcription system}
\label{ssec:transcription}

The representation format for Ancient Egyptian texts used in iClassifier is the Manuel de Codage 
\citep[MdC;][]{buurman1988mdc} transcription, which, despite some criticism \citep{nederhof2013mdc}, 
remains the standard in Egyptology. Hieroglyphic signs in MdC are represented with their Gardiner
numbers \citep[438--548]{gardiner1957grammar},\footnote{\url{https://en.wikipedia.org/wiki/Gardiner\%27s\_sign\_list}}
with additional symbols used
for denoting relative positions of signs, damaged signs, ligatures, and other information. An~example
transcription is shown in Figure~\ref{fig:mdc-example}.

Classifier signs in iClassifier are surrounded with \textasciitilde's, so the annotated version 
of the example from Figure~\ref{fig:mdc-example} is 
\texttt{U33-Z4-D21-Z1-D21-Z1-\textasciitilde D56\textasciitilde-\textasciitilde 
D54\textasciitilde}.\footnote{\url{https://thesaurus-linguae-aegyptiae.de/sentence/IBUBdWH5CJXKnkyQhOCrlBiZSCA}}
The simplest operationalisation of the classifier-identification problem is therefore seq2seq
transduction with bare transcriptions (in MdC or any other suitable scheme) as inputs and the same
encodings with tildes added when necessary as outputs. As we discuss below, however, this operationalisation
makes the transduction task unnecessarily hard for the models and considerable gains may be made
by means of some straightforward simplifications.

\section{Methods}
\label{sec:methods}

In this section, we describe our approaches to input tokenisation and output formatting (\S~\ref{ssec:preprocessing}),
the baselines (\S~\ref{ssec:baselines}), and the experimental setup (\S~\ref{ssec:experimental-setup}).

\subsection{Preprocessing}
\label{ssec:preprocessing}

The aim of the Manuel de Codage transcription system is not only to represent several hundred signs of Egyptian
hieroglyphics using numbers and Latin letters but also, as far as possible, to describe their spatial relations
in the original inscriptions since the Ancient Egyptian writing was inherently two-dimensional. Additional 
complexity comes from the ability of the transcription system to handle damaged inscriptionts, empty space, and
editorial emendations, among other things. As a result, although it is possible to
represent (a somewhat simplified version of) MdC as a context-free
grammar,\footnote{\url{https://mjn.host.cs.st-andrews.ac.uk/egyptian/res/mdc.html}} which is used, for
example, in the standard MdC-visualisation tool JSesh,\footnote{\url{http://jseshdoc.qenherkhopeshef.org/}}
this grammar is quite complex and it seems unreasonable to expect seq2seq classifiers to learn it implicitly.
Therefore we preprocessed the input by (i)~parsing it with a simplistic CFG powerful enough
to distinguish between signs, delimiters, and other elements,\footnote{The parser was implemented using
the Python package \href{https://github.com/lark-parser/lark}{Lark}. The CFG for the grammar is given in the
Appendix.} and (ii)~replacing everything except for hieroglyphs and tildes, used to mark CLFs,
with spaces.

\paragraph{Tokenisation.} The output of the previous step is a sequence of hieroglyphs in MdC,
with CLFs flanked by tildes, separated by spaces. When fine-tuning a pre-trained model with its
own tokeniser, the input must be represented as a string. If we train a model from scratch, however,
a trade-off can be made between, on one hand, longer inputs and a very small vocabulary (Latin letters, digits, and
the tilde) and, on the other hand, short inputs and a large vocabulary, where each hieroglyph from the dataset gets
its own token (784 tokens in total in our data). We call models using the small vocabulary \textit{character based} 
and models using the large vocabulary \textit{sign based}.

\paragraph{Output formatting.} Regardless of the tokenisation approach, reference outputs can be
represented in several different ways, for example:

\begin{enumerate}[noitemsep,topsep=2pt,parsep=2pt,partopsep=0pt]
    \item In the (simplified) original notation: 
    \texttt{U33 Z4 D21 Z1 D21 Z1 D56 D54} \(\rightarrow\) 
    \texttt{U33 Z4 D21 Z1 D21 Z1 \textasciitilde D56\textasciitilde\, \textasciitilde D54\textasciitilde}
    \item Without the first tilde, since each classifier in the data is unambiguously identified by a single marker:
    \texttt{U33 Z4 D21 Z1 D21 Z1 D56 D54} \(\rightarrow\) 
    \texttt{U33 Z4 D21 Z1 D21 Z1 D56\textasciitilde\, D54\textasciitilde}
    \item As a sequence of binary labels: \texttt{U33 Z4 D21 Z1 D21 Z1 D56 D54} 
    \(\rightarrow\) \texttt{0 0 0 0 0 0 1 1}
\end{enumerate}
While the first approach preserves the structure of the data, it forces the models to learn 
complicated well-formedness constraints. The second approach considerably simplifies them since the models can always first
copy the sign and then add a tilde when necessary. However, copying can still be imperfect, especially with character-based
models. The third approach completely dispenses with the original data format, but it makes enforcing
the structural constraints almost trivial. Preliminary experiments showed that resorting to binary labels gives a strong
boost in performance, and we used this approach in all reported experiments.

\subsection{Baselines}
\label{ssec:baselines}

The existence of frequent classifiers and other imbalances in the sign distribution suggest
that we may dispense with using complicated machine-learning methods altogether and predict classifiers using
sign statistics. In this study, we use the following approaches as baselines to which we compare our
sequence-to-sequence methods:

\begin{enumerate}
    \item \textbf{Top-N}: we mark \(N = 5, 10, 20, 30, 50, 100\) signs that are most-frequent classifiers
    in the training set as classifiers. \(N\) is selected using the validation set.
    \item \textbf{CLF-only}: we mark signs as classifiers if they only appear as such in the training set.
    \item \textbf{CLF-majority}: we mark signs as classifiers if they appear more frequently in this function
    in the training set.
\end{enumerate}

\subsection{Experimental setup}
\label{ssec:experimental-setup}

\paragraph{Models and training.}

We contrast the performance of sign-frequency-based baselines with three neural seq2seq models:
a character-based 3-layer encoder-decoder LSTM with a hidden dimension of 512,
a sign-based 3-layer encoder-decoder LSTM with the same hidden size, 
and ByT5-small \citep{byt5}. We thus cover both RNN-based and Transformer-based models.
Given relatively short input lengths, we keep RNNs simple and do not equip them with attention.

Importantly, the small version of ByT5 is still a considerably larger model compared to the seq2seq LSTMs
and therefore harder to train on a small dataset. However, there is a possibility that its extensive
pre-training on data from other languages gives it enough inductive bias to tackle a novel language,
even with a non-orthodox transcription. 

The batch size and learning rate for the models reported below were selected using grid search
on the development set, and the models were trained until there was no improvement on the 
development set for 5 epochs.\footnote{The code and the dataset used for the analyses are available
at \url{https://git.sr.ht/~macleginn/ml4al-iclassifier-paper-code/tree}}

\paragraph{Evaluation metric.} As the evaluation metric, we use the average number of
mistakenly classified signs in the test-set data points.

More precisely, we split the output
of the decoder on whitespaces, pad the resulting vector of labels with zeros if it is too short,
and convert any non-1 elements to zeros as well. This corresponds to a conservative procedure that,
given an input sequence of signs, outputs a sequence of signs with marked classifiers 
and without NAs, which is how the system is arguably supposed to work in practice.

\section{Results}
\label{sec:results}

\begin{table}[t]
\centering
\begin{tabular}{@{}lrrr@{}}
\toprule
Model & Dev & Test & Narratives \\ \midrule
CLF only & 1.23 & 1.23 & 1.39 \\
Top-50 CLF & 0.46 & 0.47 & 1.07 \\
CLF majority & 0.27 & 0.28 & 0.49 \\
LSTM (char) & 0.2 & 0.21 & 3.07 \\
LSTM (sign) & 0.14 & 0.11 & 0.38 \\
ByT5 small & 0.08 & \textbf{0.1} & \textbf{0.35}\\ \bottomrule
\end{tabular}
\caption{Average number of misclassified signs per data point on the Coffin Texts corpus (dev and test) and
the Late Egyptian narratives (out-of-domain).
\textbf{CLF only}: signs only found as CLFs in the training set are marked as CLFs.
\textbf{Top-50 CLF}: 50 signs that are most frequently found as CLFs marked as CLFs. 
\textbf{CLF majority}: signs that are more frequently found as CLFs than as regular signs marked as CLFs. 
\textbf{LSTM (char)}: character-based 3-layer encoder-decoder model with the hidden dimension of 512.
\textbf{LSTM (sign)}: sign-based 3-layer encoder-decoder with the hidden dimension of 512.}
\label{tab:results}
\end{table}

The performance of the trained models on the development and test subsets of the Coffin Texts corpus
and on the out-of-domain (OOD) data from the Late Egyptian corpus is reported in Table~\ref{tab:results}.
Several observations can be made.

First, the Coffin Texts are shown to be quite homogeneous: the performance drop between the development
and test sets is marginal, with one model (sign-based seq2seq LSTM) even gaining 3 performance percentage
points.

Secondly, the character-based LSTM model does not perform well: it barely beats the CLF-majority baseline
and suffers performance collapse on the OOD data. The sign-based LSTM, on the other hand,
is very competitive, even on the OOD test dataset, where, unlike ByT5, it had to contend with UNK
tokens, mapped to SOS tokens.

Thirdly, ByT5, despite not being trained on any directly comparable data and being character based, beats
the sign-based seq2seq LSTM model both on the in-domain and on the out-of-domain test sets. This suggests that
there may be a decent possibility for knowledge transfer between classifier languages.

Finally, the CLF-majority baseline, despite its conceptual simplicity, demonstrates tolerable performance
and with some additional tuning may be used as a lightweight method that can dynamically respond
as new data points are added.

It must be pointed out that the array of possible CLFs is very wide, given the existence of phonetic classifiers. 
Despite the homogeneity of the Coffin Texts data, the test set contains 
19 CLFs not found in either test or dev subsets; 17 of them are only used once. Conversely, 156 CLFs were 
encountered only once in the combined test and dev set. The OOD test set, despite being twice smaller than 
the in-domain one, also has 13 new CLFs. This does not preclude the possibility of ever identifying such 
classifiers (human expert annotators can do this by, e.g., analysing the structure of different lexical 
items across contexts), but this considerably raises the demands on the size of the training set.


\section{Conclusion}
\label{sec:conclusion}

\begin{table}[t]
\begin{tabular}{@{}lllllllll@{}}
\toprule
0    & 1    & 2    & 3   & 4   & 5  & 6 & 7 & 8 \\ 
1403 & 4113 & 2195 & 573 & 112 & 20 & 6 & 0 & 1 \\ \bottomrule
\end{tabular}
\caption{Counts of data points with different number of CLFs in the train and
dev subsets of the Coffin Texts dataset.}
\label{tab:clf-counts}
\end{table}

This study is a first step towards creating a trained system for identification and analysis
of classifiers and other sign functions in ancient complex scripts. 
It demonstrates that it is possible to achieve respectable
error rates on this task on in-domain data, with \(\approx 0.1\) mistakenly identified classifiers per data point.
Given a high number of data points with several classifiers (cf.\ Table~\ref{tab:clf-counts}),
this translates to correct analysis of most wordforms. The accuracy falls significantly on out-of-domain data,
but it must be noted that our OOD test set is distinguished from the training set not only by a different
genre (narratives vs.\ religious texts) but also by at least 400 years of language evolution.

Future work, in addition to improving model accuracy, could be directed toward providing a more
fine-grained classification of sign functions by leveraging the distinction between semantic
and grammatical classifiers and phono-repeaters.

\section*{Acknowledgments}

The research for and preparation of this article were supported by ISF grant number 1704/22, 
\enquote{Exploring the minds of Ancient Egypt and Ancient China~-- a comparative network analysis of the classifier 
systems of the scripts}, and ISF grant number 735/17, \enquote{Classifying the other: The classification of Semitic 
loanwords in the Egyptian script}, awarded to ArchaeoMind Lab, The Hebrew University Jerusalem (\url{www.archaeomind.net}), 
PI Orly Goldwasser. 


Jorke Grotenhuis' work was supported by the Israel Academy of Sciences and Humanities \& Council for Higher 
Education Excellence Fellowship Program for International Postdoctoral Researchers and the ArchaeoMind Lab. 


\bibliography{main}


\appendix

\section*{Appendix: A~CFG for parsing MdC}
\label{sec:appendix}

In Lark notation:

{\small
\begin{verbatim}
token : sequence (delimiters sequence)*

delimiters : delimiter+

sequence : left_paren sequence right_paren
    | tilde sequence tilde
    | sequence delimiters sequence
    | classified_sign

left_paren : "("
right_paren : ")"

classified_sign : code suffix?
    | tilde code tilde suffix?

suffix : ligature_pos
    | damage
    | ligature_pos damage
    | damage ligature_pos

code : /[a-zA-Z]+[0-9]*[a-zA-Z]*/
    | /[0-9]+/
    | "#b-..#e"
    | "#b"
    | "#e"
    | "[&"
    | "&]"
    | "."

damage : /#\d+/

ligature_pos : /\{\{\d+,\d+,\d+\}\}/

delimiter : "-"
    | ":"
    | "\\"
    | "\\\\"
    | "\\\\\\\\"
    | "_GROUPING_"
    | "^"
    | "("
    | ")"
    | "&"
    | "{"
    | "}"
    | ","
    | "*"
    | "_"

tilde : "~"
\end{verbatim}
}

\end{document}